\documentclass[runningheads]{llncs}
\usepackage{graphicx}
\usepackage{multirow}
\usepackage{bbding}
\usepackage{color}
\usepackage{booktabs}
\usepackage{hyperref}
\usepackage[misc]{ifsym}
\begin{document}
\title{Spatial Attention and Syntax Rule Enhanced Tree Decoder for Offline Handwritten Mathematical Expression Recognition}
\titlerunning{SS-TD}
%
\author{Zihao Lin\inst{1} \and
Jinrong Li\inst{2} \and
Fan Yang\inst{1}\and
 Shuangping Huang\inst{1}\inst{3}\Letter \and
Xu Yang\inst{4}\Letter\and \\
Jianmin Lin\inst{2}\and
Ming Yang\inst{2}
}

\authorrunning{Z.Lin et al.}
%
\institute{South China University of Technology, Guangzhou, China
\email{linzihao2637@gmail.com, maxgundam@hotmail.com, eehsp@scut.edu.cn}
\and
CVTE Research, Guangzhou, China\\
\email{\{lijinrong, linjianmin, yangming\}@cvte.com}
 \and
Pazhou Lab, Guangzhou, 510330, China\\
\email{eehsp@scut.edu.cn}
\and
GRGBanking Equipment Co. Ltd. Guangzhou, China\\
\email{yxu8@grgbanking.com}}

%
%
\maketitle              
\begin{abstract}
Offline Handwritten Mathematical Expression Recognition (HMER) has been dramatically advanced recently by employing tree decoders as part of the encoder-decoder method. Despite the tree decoder-based methods regard the expressions as a \LaTeX{} tree and parse 2D spatial structure to the tree nodes sequence, the performance of existing works is still poor due to the inevitable tree nodes prediction errors. Besides, they lack syntax rules to regulate the output of expressions. In this paper, we propose a novel model called \textbf{S}patial Attention and \textbf{S}yntax Rule Enhanced \textbf{T}ree \textbf{D}ecoder (SS-TD), which is equipped with spatial attention mechanism to alleviate the prediction error of tree structure and use syntax masks (obtained from the transformation of syntax rules) to constrain the occurrence of ungrammatical mathematical expression. In this way, our model can effectively describe tree structure and increase the accuracy of output expression. Experiments show that SS-TD achieves better recognition performance than prior models on CROHME 14/16/19 datasets, demonstrating the 
effectiveness of our model. 

\end{abstract}
\section{Introduction}
Offline handwritten mathematical expression recognition (HMER) has many applications like searching for questions in education, recording payment in finance, and many other fields. However, it is a challenge to recognize handwritten mathematical expression (HME) due to ambiguity of handwritten symbols and complexity of spatial structures in the expression. It means HMER should not only correctly recognize the symbols but also analyze the relationship between them, which put forward great difficulties in the recognition performance.

With the rapid advancement of deep learning-based methods for HMER, some studies have proposed string decoder~\cite{DBLP:conf/icfhr/LiJLZ20,DBLP:conf/icfhr/TruongNPN20,DBLP:conf/iiki/WangSW18,DBLP:journals/ijcv/WuYZZL20,DBLP:journals/pr/ZhangDZLHHWD17}~with handwritten mathematical expression images as input and LaTeX strings of expression as predicted targets. However, their works emphasize symbol recognition but seldom consider structure information, which leads to the unsatisfactory result in the recognition of expressions with complex structures like $\sqrt{2+\sqrt{2}}$ .

Moreover, in recent years, research based on tree decoder~\cite{DBLP:journals/corr/abs-2203-01601,DBLP:journals/tmm/ZhangDYSD21,DBLP:conf/icml/ZhangDYSW020}~which focuses on the structure was emerging in the field of HMER. The methods based on tree decoder inherently represent expression as tree structures which is more natural for HMER. Among them, Zhang et al.~\cite{DBLP:conf/icml/ZhangDYSW020}~proposed a tree decoder (DenseWAP-TD) to predict the parent-child relationship of trees during decoding procedure, which achieved the best recognition performance at that time. Although the tree decoder models the tree structure of HMEs explicitly, the structure prediction accuracy may decrease as they only use the node symbol information to predict the node structure. What's more, as shown in Fig.1(c), the previous methods output an ungrammatical expression in which the symbol `2' can not have a `Subscript' relation with the symbol `$\sum$'. It means the previous tree decoders may lack the syntactic rules to generate the grammatical expressions. 
\begin{figure}[bth]
\center{\includegraphics[width=0.7\textwidth]{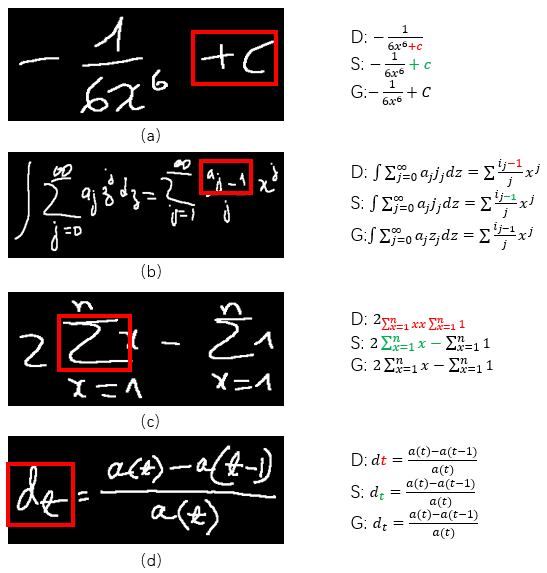} }   
\caption{Comparison with the previous method DenseWAP-TD. `D',`S',`G' stand for DenseWAP-TD, SS-TD and Ground-Truth. (a) and (b) denote parent node prediction error of previous method. (c) and (d) denote the relation prediction error of previous method. Symbols in red are the prediction errors. Symbols in green is the corresponding prediction} \label{fig1}
\end{figure}

In order to solve the above problems, inspired by the DenseWAP-TD model~\cite{DBLP:conf/icml/ZhangDYSW020}~, we proposed a novel tree decoder (SS-TD) integrating spatial attention and syntax rules which is shown in Fig.2. First, we introduce spatial attention mechanism to predict the parent node, which effectively improves the structure accuracy. What’s more, for purpose of reducing the generation of ungrammatical mathematical expressions, we transform the grammar of the relationship between expression symbols into the syntax mask and introduce it into the process of relation prediction. The comparison results shown in Fig.1 illustrate that our model can deal with the problems of the previous method. To further confirm the effectiveness of our model, we conduct the experiments on CROHME dataset. The results show that our model consistently achieves higher recognition rates over many other methods, demonstrating the effectiveness of our model for offline HMER.

The main contributions of this paper are as follows:

$\bullet$ We proposed a new tree decoder (SS-TD) which can effectively adapt encoders of other methods to achieve higher recognition performance.

$\bullet$ To improve the structure accuracy, we introduce spatial attention enhancement to predict node information. Besides, we introduce syntax masks to constrain the generation of ungrammatical mathematical expressions.

$\bullet$ We demonstrate the advantage of SS-TD on the CROHME14/16/19 dataset by experiments and achieve very competitive results.

\section{Related Works}
Early traditional methods~\cite{cortes1995support,DBLP:conf/icdar/OkamotoIT01,DBLP:journals/tip/QianH96}~employed sequential approaches, which implemented symbol recognition and structural analysis separately. Recognition errors might be accumulated constantly in these methods. On the other hand, some studies~\cite{alvaro2016integrated,DBLP:journals/prl/AwalMV14,DBLP:conf/icpr/ChanY98}~handled it as a global optimization to settle the problem of error accumulation, but made the process more inefficient.

Recently, deep learning-based encoder-decoder approaches were investigated by many studies on offline HMER. Some of them used string decoder to generate LaTeX strings. WAP~\cite{DBLP:journals/pr/ZhangDZLHHWD17}~proposed by Zhang et al. employs a fully convolutional network encoder and an attention-equipped decoder, which achieved state-of-the-art at that time on CROHME 14/16 datasets. Then, Zhang et al.~\cite{DBLP:conf/icpr/ZhangDD18}~proposed a DenseNet encoder to improve the WAP, which is also employed in our model. Wu et al.~\cite{DBLP:journals/ijcv/WuYZZL20}~proposed an adversarial learning strategy to overcome the difficulty about writing style variation. However, these studies lack structure relationship awareness, which leads to inevitable errors in some complex mathematical expression recognition.

Some works~\cite{DBLP:conf/iclr/Alvarez-MelisJ17,DBLP:journals/corr/abs-1810-00314,DBLP:conf/nips/ChenLS18,DBLP:conf/naacl/DyerKBS16,DBLP:journals/corr/abs-1908-00449}~employed tree decoder for various tasks, which is more natural to represent the structure. In this regard, some studies employed tree decoders to model the structure of mathematical expression. For online HMER, Zhang et al. proposed a tree decoder method SRD~\cite{DBLP:journals/tmm/ZhangDYSD21}~to generate tree sequence of expressions. Wu et al.~\cite{DBLP:conf/AAAI/WuJW21} treated the problem as a graph-to-graph learning problem, which essential is the same as the tree decoder. On the other hand, for offline HMER, Zhang et al. proposed a tree-structured decoder (DenseWAP-TD), which decomposes the structure tree into a sub-tree sequence to predict the parent-child nodes and the relationship between them. However, the existing methods based on tree decoder attempt to take syntactic information into account, which still could not guarantee the grammatical accuracy of the output expression. Yuan et al.~\cite{DBLP:journals/corr/abs-2203-01601}~incorporated syntax information into an encoder-decoder network, which gave us a lot of inspiration.

As described above, although some tree decoders have been proposed, there is still a great challenge on how to further improve the accuracy in offline HMER and effectively combine semantic information into the network. For the purpose of solving these problems, we integrate spatial attention mechanism and syntax rules into the tree decoder to deal with complex mathematical expressions and adapt to the syntactic rationality of expressions as much as possible.

\section{Proposed Method}
Our model takes the input HME images and yields the symbol tree structure as shown in Fig.2. We adopt an encoder to extract the feature vector sequence of the expression images. The decoder takes the feature sequence as input and then yields sequential triples step by step. Each triple which represents the sub-tree structure includes a child node, a parent node, and the relationship between parent-child nodes, represented as ($y_t^c,y_t^p,y_t^{rel}$). The child node $y_t^c$ is composed of the predicted symbol and its tree sequence order. The predicted parent node $y_t^p$ is one of the previous child nodes $y_{1,2...,t-1}^c$ which has best match with current child node $y_t^c$ in the semantic and spatial information. The connection relationship $y_t^{rel}$ between current parent and child nodes is represented as 6 forms: Above, Below, Sup, Sub, inside, and Right.
\begin{figure}[bth]
\includegraphics[width=\textwidth]{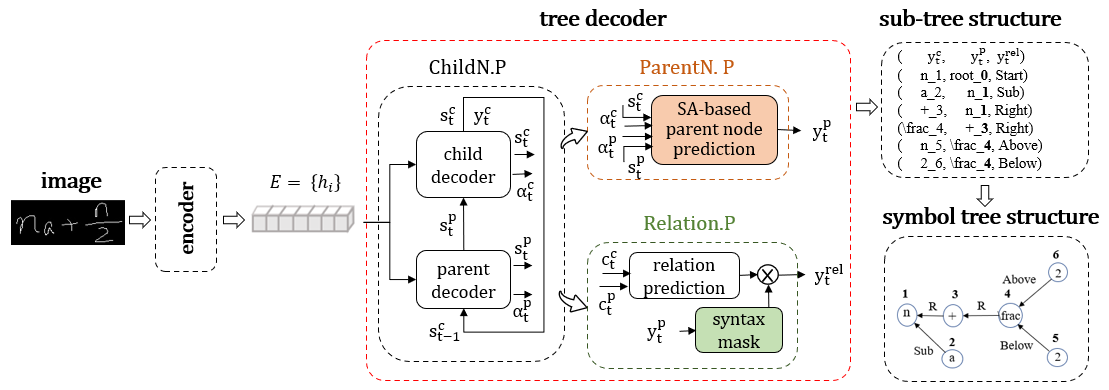}
\caption{Overview of SS-TD. The encoder is a DenseNet. `SA-based parent node prediction' denotes the parent node prediction module equipped spatial attention (SA). `Syntax mask' denotes transformed syntax rules introduced in our model. The details of the tree decoder are illustrated in Fig.3}\label{fig2}
\end{figure}
\begin{figure}[bth]
\center{\includegraphics[width=0.8\textwidth]{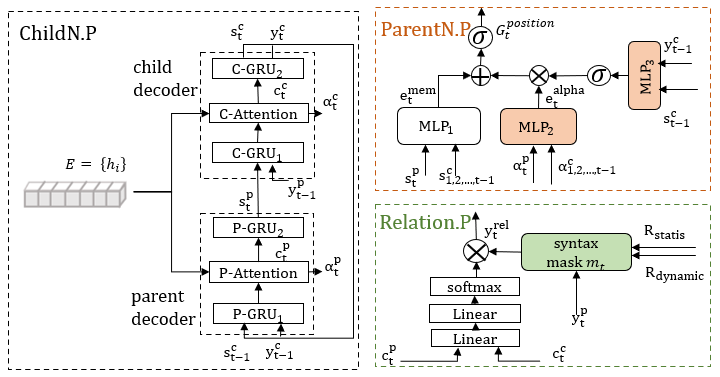}}
\caption{The main prediction module in the decoder of SS-TD. `ChildN.P' represents the child node prediction module. `ParentN.P' represents the parent node prediction module. `Relation.P' represents the relation prediction module.} \label{fig3}
\end{figure}

Just like the previous methods based on encoder-decoder~\cite{DBLP:conf/icpr/ZhangDD18}~, we also employ DenseNet to  encode the HME images $I\in R^{H\times W\times C}$ and extract the feature vector sequence $E=\left\{ h_i \right\} \in R^{H'\times W'\times D}$, where $h_i\in R^{D}$.
Then, the feature sequence is input into the child decoder and the parent decoder respectively to recognize the symbol step by step. For the purpose of modeling the expression structure, the decoder of SS-TD consists of three prediction modules: 1)the child node prediction module shown as ChildN.P in Fig.2 and Fig.3 to predict the current child node information $y_t^c$; 2)the parent node prediction module shown as ParentN.P in Fig.2 and Fig.3 to predict the parent node $y_t^p$ from the previous child nodes which has best match with current child node;
3)the relationship prediction module shown as Relation.P in Fig.2 and Fig.3 to predict the relationships $y_t^{rel}$ between parent and child nodes which are mentioned above. Consequently, the prediction modules yield the triples which can be transformed to the symbol tree structure. At last, we traverse the tree structure to obtain the \LaTeX{} strings. 
\subsection{Child Node Prediction Module}
Like the previous tree decoder-based method DenseWAP-TD, we also adopt a parent decoder and a child decoder to recognize the child node symbol in sequence as shown in Fig.3. These two decoders generate a tuple including the symbol and its recognition order, which is defined as child node information.  

The parent decoder consists of two Gated Recurrent Units ($GRU$) layers and an attention block as shown in the left part of Fig.3. It takes the previous child node $y_{t-1}^c$ and its hidden state $s_{t-1}^c$ as input, and outputs the parent context vector $c_t^p$ and its hidden state $s_t^p$, which is also the input of the child decoder and the other two modules ParentN.P and Relation.P which shown in the right of Fig.3:
\begin{equation}
c_t^p=\sum_{i=1}^L a_{ti}^p h_i
\end{equation}
\begin{equation}
s_t^p={GRU}_2^p(c_{t}^p,\hat{s}_{t}^p)
\end{equation}
where $h_i$ is the i-th element of feature map $E$. $L=H'\times W'$. $s_t^p$ is the prediction of current parent hidden state. $a_t^p=\left\{ a_{ti}^p\right\}$ is the current parent attention probabilities which is computed as follows: 
\begin{equation}
\hat{s}_t^p={GRU}_1^p(y_{t-1}^c,s_{t-1}^p)
\end{equation}
\begin{equation}
a_t^p=f_{att}^p(\hat{s}_t^p,a_{1,2...,t-1}^p)
\end{equation}
where $f_{att}^p$ represents attention function followed the DenseWAP-TD~\cite{DBLP:conf/icml/ZhangDYSW020}~. 

The structure of child decoder is almost the same as the parent decoder. The decoder inputs the previous parent node $y_{t-1}^p$ and its hidden state $s_t^p$, and outputs the child context vector $c_t^c$ and its hidden state $s_t^c$. 

We take the previous parent node $y_{t-1}^p$, the child node hidden state $s_t^c$ and its context vector $c_t^c$ as input to compute the probability of each output predicted child node $p(y_t^c)$:
\begin{equation}
p(y_t^c)=softmax(W_{out}^c(W_ey_{t-1}^p+W_hs_t^c+W_cc_t^c))
\end{equation}
where $W_{out}^c\in R^{S}$ is a full connection layer parameter. $S$ is the size of the recognition symbol set.

The classification loss of child node prediction module is:
\begin{equation}
L_c=-\sum_{t=1}^Tlog(p(y_t^c))
\end{equation}

During the test process, we use the previous one-hot vector of the parent decoder and the child decoder in greedy algorithm to predict the child node:
\begin{equation}
\hat{y}_t^c=\mathop{\arg\max}\limits_{y^c}\ p(y_t^c)
\end{equation}
where $y^c$ represents all the symbols in the symbol set. 

\subsection{Spatial Attention-based Parent Node Prediction Module}
We treat predicting the parent node as finding previous child nodes which have the best match with current child node. We define the order of previous child nodes as the parent node position $\hat{y}^{pos}$, which is the predicted target in this module.

We first use the semantic information representing the hidden state of the nodes to predict the parent nodes. We employ Multilayer Perceptron (MLP) architecture to obtain the semantic energy factor $e_{ti}^{mem}$ which is shown as the $\rm MLP_1$ block in Fig.3:
\begin{equation}
e_{ti}^{mem}=v_{mem}^T\tanh(W_{mem}s_t^p+U_{mem}s_{1,2...,t-1}^c)
\end{equation}
where $e_{ti}^{mem}$ is the semantic energy of $i-th$ element at decoding step $t$. $s_t^p$ is the hidden state of parent decoder. $s_{1...t-1}^c$ are the hidden states of child nodes in history. 

However, only using semantic information to predict the parent node may lead to inevitable error especially when facing the identical symbols of expression. Thus, we introduce spatial attention mechanism to alleviate the prediction error of parent node.

The spatial information of the parent node and previous child nodes is represented as the attention distribution $a_t^p$ and $a_{1,2...,t-1}^c$ respectively. The spatial energy factor is computed as follows:
\begin{equation}
e_{ti}^{alpha}=v_{alpha}^T\tanh(W_{alpha}p(a_t^p)+U_{alpha}p(a_{1,2...,t-1}^c))
\end{equation}
where $e_{ti}^{alpha}$ is the spatial energy of $i-th$ element at decoding step $t$. $p(\cdot)$ represents
adaptive pooling. The size of its output vector is 4$\times$32. 

We further employ the gate mechanism to control the input of updated spatial position information as shown in the ParentN.P block of Fig.3. The gate mechanism is computed as follows:
\begin{equation}
g_t=\sigma (W_{yg}y_{t-1}^c+U_{sg}s_{t-1}^c)
\end{equation}
\begin{equation}
e_{ti}^{position}=e_{ti}^{mem}+g_t\odot e_{ti}^{alpha}
\end{equation}
\begin{equation}
G_{ti}^{position}=\sigma e_{ti}^{position}
\end{equation}
where $\odot$ means the element-wise product.  

The loss of the parent node prediction module is defined as a cross entropy loss:
\begin{equation}
L_{pos}=-\sum_{t-1}^T\sum_{i=1}^L[\bar{G}_{ti}^{position}log(G_{ti}^{position})+(1-\bar{G}_{ti}^{position})log(1-G_{ti}^{position})]
\end{equation}
where $\bar{G}_{ti}^{position}$ represent the ground-truth of the parent node. If $i-th$ child node in history is the current parent node, $\bar{G}_{ti}^{position}$ is 1, otherwise 0.

In the testing stage, we compute the parent node position $\hat{y}^{pos}$:
\begin{equation}
\hat{y}^{pos}=\mathop{\arg\max}\limits_{g}\ G^{position}
\end{equation}
\indent And then the prediction of parent node $\hat{y}^p$ is obtained by using the $\hat{y}^{pos}$ which also represents the child node order. After that, we put it into the child decoder to predict the child node.

\subsection{Syntax Rule-based Relation Prediction Module}
In order to predict the grammatical expression, we introduce syntax rules into the relation prediction module. Two crucial syntax rules related to conjunction relations are summarized as follows:

$\bullet$ Different mathematical symbols have different restrictions of connection relation, like the `$\sum$' can not have the `Inside' relation with other symbols. 

$\bullet$ Each mathematical symbol cannot have repeated connection relations at the same time, like one symbol can not have 'Right' relation with two other symbols.
\begin{table}[bth]
\caption{Static syntax mask. The connection relationships of the symbol are transformed into the Static syntax mask, which is designed in the sequence of [Right, Sup, Sub, Above, Below, Inside]}\label{tab1}
\begin{center}
    \setlength{\tabcolsep}{1.5mm}
    \renewcommand\arraystretch{1.3}
    \begin{tabular}{|c|cccccc|l|}
\hline
 &  Right&Sup&Sub&Above&Below&Inside & Symbol\\
\hline
\begin{minipage}[b]{0.04\columnwidth}
		\centering
		\raisebox{-.4\height}{\includegraphics[width=\linewidth]{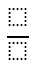}}
	\end{minipage}
&  1&1&0&1& 1& 0 & $\setminus$frac\\
\hline
\begin{minipage}[b]{0.04\columnwidth}
		\centering
		\raisebox{-.4\height}{\includegraphics[width=\linewidth]{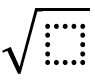}}
	\end{minipage}
	&  1& 0 &0 &0 &0 &1 & $\setminus$sqrt\\
\hline
$\rm \Sigma$, $\rm \Pi$ & 1 &1& 1 &1 &1 &0 & [$\setminus$sum, $\setminus$prod]\\
\hline
\begin{minipage}[b]{0.08\columnwidth}
		\centering
		\raisebox{-.4\height}{\includegraphics[width=\linewidth]{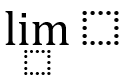}}
	\end{minipage}& 1& 0 &0 &0 &1 &0  & $\setminus$lim\\
\hline
Letter & 1 &1 &1 &0& 0& 0  & \begin{tabular}[c]{@{}l@{}}the Greek letter,\\English letter,\\ the ending symbols...\end{tabular}\\
\hline
Number & 1 &1 &0 &0 &0& 0  & \begin{tabular}[c]{@{}l@{}}number,\\ Trigonometric functions\end{tabular}\\
\hline
Bin Symbol & 1& 0 &0& 0 &0& 0 & \begin{tabular}[c]{@{}l@{}}the beginning symbols,\\ operator...\end{tabular}\\
\hline
\end{tabular}
\end{center}
\end{table}

Then, to integrate these rules into the model training process, we creatively put forward the syntax mask representing the mathematical connection relationship. According to the two syntax rules, we define the syntax mask of mathematical symbols. If the mathematical symbols can be connected to its related symbol in a relation of Right, Sup, Sub, Above, Below, Inside, the corresponding syntax mask value is 1, otherwise 0. 

For the first syntax rule which is regarded as the prior knowledge, the mask of it could be obtained in advance as shown in Table.1. For example, the mask of the symbol `$+$' is `100000' since it only has one relation `Right'. Besides, we group some symbols which have the same syntax mask, such as the Letter as shown in Table 1. In addition, we deal with the symbols which can be represented as different syntax masks by the logical `OR' operator. For example, the mask of symbol `e' is `111000' when it represents the lowercase letter, while the mask is `110000' when it represents the irrational constant. To sum up, the mask of `e' is `111000$\parallel$110000=111000', where `$\parallel$' is the logical `OR' operator.

We present the static syntax mask matrix as $R_{static}\in R^{C\times 6}$, and the static syntax mask of parent node as $m_t^s\in R^6$:
\begin{equation}
m_t^s = y_t^pR_{static}
\end{equation}
where $y_t^p$ is the one-hot vector of the symbol in the current parent node.

On the other hand, for the second syntax rule, we need to acquire information about the previous relations between parent-child nodes. To solve the problem, we propose a dynamic mask matrix $R_{dynamic}^t$ which is initialized into an all-zero matrix to store the historical relations in step $t$. For example, when the parent node and its relationship are predicted to be `$\setminus$sqrt' and `Inside', the value of the corresponding row of `$\setminus$sqrt' and the corresponding column of `Inside' in $R_{dynamic}^t$ should be updated as `1'. The dynamic syntax mask of parent node is represented as $m_t^d\in R^6$:
\begin{equation}
m_t^d = y_t^pR_{dynamic}^{t-1}
\end{equation}
\begin{equation}
m_t = m_t^s\otimes m_t^d
\end{equation}
where $\otimes$ is the logical exclusive or operator. The update of dynamic mask matrix $R_{dynamic}^t$ is computed as follows:
\begin{equation}
R_{dynamic}^t = R_{dynamic}^{t-1} + {y_t^p}^Ty_t^{rel}
\end{equation}
where ${y_t^p}^T$ represents the transposition of the parent node $y_t^p$.

As shown in Fig.3, We compute the probabilities of the relation prediction $p(y_t^{rel})$ as follows:
\begin{equation}
p(y_t^{rel})=softmax(m_t\odot (W_{out}^{rel}(W_{cp}c_t^p+W_{cc}c_t^c)))
\end{equation}
where $\odot$ represents multiplication element by element.

The classification loss of relation prediction module is computed as follows:
\begin{equation}
L_{rel}=-\sum_{t=1}^Tlog(p(y_t^{rel}))
\end{equation}
\indent In the testing stage, we take the relation of maximum probability as final prediction relation:
\begin{equation}
\hat{y}^{rel}=\mathop{\arg\max}\limits_{y^{rel}}\ p(y_t^{rel})
\end{equation}
where $y^{rel}$ represent the relations set.

\subsection{Total Loss}
We use an attention self-regularization loss to speed up network convergence. Specifically, a Kullback-Leibler divergence is employed to measure the difference in the distribution of attention generated by parent and child decoder:
\begin{equation}
L_{alpha}=\sum_{t=1}^T\hat{a}_t^plog(\frac{\hat{a}_t^p}{a_t^p})
\end{equation}

Thus, the training objective of the SS-TD is to minimize the loss as follows:
\begin{equation}
L=\lambda_1L_c+\lambda_2L_{pos}+\lambda_3L_{rel}+\lambda_4L_{alpha}
\end{equation}
where, $\lambda$ is the weight of each term.

\section{Experiments}
\subsection{Dataset}
We verified our proposed model on the CROHME dataset~\cite{DBLP:conf/icdar/MahdaviZMVG19,DBLP:conf/icfhr/MouchereVZG16,DBLP:conf/ijdar/MouchereZGV16}~ which is the most widely used for HMER. The CROHME training set contains 8835 HMEs, 101 math symbol classes. There are 6 common spatial math relations (Above, Below, Right, Inside, Superscript and Subscript) in our implementation. We evaluate our model on CROHME
14/16/19 test set. Among them, the CROHME 2014 test set contains 986 HMEs, which is evaluated by most advanced model. And CROHME 2016 and 2019 test sets are collected and labeled for research after that, which contain 1147 and 1119 HMEs.
\subsection{Implementation Details}
For the training loss, we set $\lambda_1=\lambda_2=\lambda_3=1$ in our experiment indicating the same importance for the modules, and set $\lambda_4=0.1$.

Following the DenseNet-WAP~\cite{DBLP:conf/icpr/ZhangDD18}~, We employ DenseNet as the encoder, which is composed of three DenseBlocks and two Transition layers. Each DenseBlock contains 48 convolution layers. The convolution kernel size is set to $11\times11$ for computing the coverage vector.

Our model follows the previous tree decoder-based method DenseWAP-TD using two decoders to recognize the symbols, which has redundant parameters. We first simplify the parameters to optimize the model, where the dimensions of child attention, parent attention and memory attention are set to 128 instead of 512 and the embedding dimensions of both child node and parent node are set to 128 instead of 256. As shown in the second row of Table 2, the parameters are reduced by near half and the inference speed increases by more than 50 ms without affecting the recognition performance. Regarding this, we take the same parameters into SS-TD to pursue better performance. Due to the introduced syntax masks, the complexity of our model is quite more than the simplified DenseWAP-TD, in which the gap is still smaller than that between DenseWAP-TD and simplified DenseWAP-TD.
\begin{table}[bth]\centering
\caption{Performance on CROHME 2014 of our model versus DenseWAP-TD and Simplified DenseWAP on Expression Rate (ExpRate) (in $\rm \%$). Inference speed is represented as Speed. The number of parameters is shown in the last column.}\label{tab2}
\begin{center}
    \setlength{\tabcolsep}{1.5mm}
    \renewcommand\arraystretch{1.3}
\begin{tabular}{cccc}
\hline
Model &  $\rm Exp.Rate_{latex}(\%)$ & Speed & Params\\
\hline
DenseWAP-TD & 49.10  & 144.9 & 7.93M\\
Simplified DenseWAP-TD &  49.32 & \textbf{92.56} & \textbf{4.68M}\\
\textbf{SS-TD} & \textbf{52.48} & 94.48 & 4.77M\\
\hline
\end{tabular}
\end{center}
\end{table}

We utilized the Adadelta algorithm [29] for optimization and conduct on experiments. Besides, the framework was implemented in Pytorch. Experiments were conducted on 2 NVIDIA GeForce RTX 3090.
\subsection{Ablation Experiment}
In order to verify the performance improvement of SS-TD brought by the integration of spatial attention and syntax rules, we conduct ablation experiments on CROHME 14 datasets.
\begin{table}[bth]\centering
\caption{Results of Ablation Experiment in $\rm \%$ on CROHME 2014. `$\rm Exp.Rate_{tree}$' denotes the expression recognition rate of tree structure. `$\rm Exp.Rate_{latex}$' denotes the expression recognition rate of LaTeX string. `$\rm WER_{pos}$' denotes the word error rate of parent node prediction. `$\rm WER_{rel}$' denotes the word error rate of predicted relationship between parent-child nodes.}\label{tab3}
\begin{center}
    \setlength{\tabcolsep}{1.2mm}
    \renewcommand\arraystretch{1.3}
\begin{tabular}{c|c|c|cccc}
\hline
\multirow{3}{*}{\begin{tabular}[c]{@{}l@{}}Spatial\\ information\end{tabular}} &  \multirow{3}{*}{\begin{tabular}[c]{@{}l@{}}Static\\ syntax \\mask\end{tabular}} & \multirow{3}{*}{\begin{tabular}[c]{@{}l@{}}Dynamic\\ syntax \\mask\end{tabular}} & \multicolumn{4}{c}{CROHME 2014}\\
&&&&&&\\
 &  & & $\rm Exp.Rate_{tree}\uparrow$ & $\rm Exp.Rate_{latex}\uparrow$& $\rm WER_{pos}\downarrow$& $\rm WER_{rel}\downarrow$\\\hline
 &  & & 49.93 & 50.14& 8.64 & 6.53\\
$\surd$ &  & & 51.47& 51.47 & 7.12 &6.23\\
$\surd$& $\surd$ & &52.02 &52.10 &6.98 &5.56\\
$\surd$& $\surd$ & $\surd$ & \textbf{52.48}& \textbf{52.48} & \textbf{6.72}& \textbf{5.12}\\
\hline
\end{tabular}
\end{center}
\end{table}
\begin{figure}[bth]
\center{\includegraphics[width=0.9\textwidth]{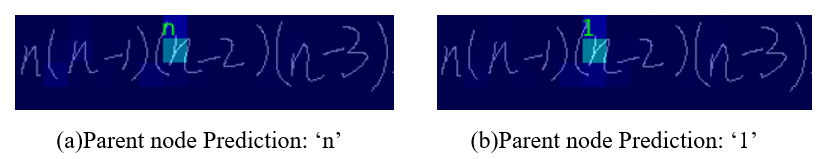}}
\caption{The visualization of spatial attention distribution. (a) denotes the correct parent node prediction by integrating spatial information. (b) denotes the error prediction by only using semantic information. Blocks in blue denote the attention distribution.} \label{fig5}
\end{figure}

As shown in Table 3, after we introduce spatial attention mechanism, the $\rm Exp.Rate_{tree}$ increases by 1.54$\%$ and the $\rm WER_{pos}$ decreases by 1.52$\rm \%$. To further prove the improvement is due to the spatial attention mechanism itself instead of the attention parameters, we compare the precision($\%$)$/$Params(M) of attention employed and attention not employed, in which the value of attention employed is 10.79 and 0.12 higher than attention not employed. In addition, as shown in Fig.4, the result predicted by only using semantic information is the symbol `1' which is incorrect obviously. And the attention distribution of child node symbols correctly indicate the right symbol `n'. Hence the spatial attention enhancement achieved improvements and this demonstrates the importance of spatial information for the parent node position.

On the other hand, as illustrated in the third row and the forth row of Table 3, the performance has a great improvement on the whole after adding static and dynamic syntax masks. It is worth noting that the gap between $\rm Exp.Rate_{tree}$ and $\rm Exp.Rate_{latex}$ becomes smaller, which indicates the grammar accuracy of the output tree structures has great improvement. 
In general, after the enhancement we introduce into the model, $\rm Exp.Rate_{tree}$ and $\rm Exp.Rate_{latex}$ increase by 2.55$\%$ and 2.34$\%$, $\rm WER_{pos}$ and $\rm WER_{pos}$ decrease by 1.92$\%$ and 1.41$\%$, respectively, which validates the effectiveness of the spatial attention and syntax rules.

\subsection{Performance Comparison}
\begin{table}[bth]\centering
\caption{Evaluation of offline HMER on CROHME 14/16/19 ($\rm \%$)}\label{tab4}
\begin{center}
    \setlength{\tabcolsep}{1.5mm}
    \renewcommand\arraystretch{1.3}
\begin{tabular}{l|c|c|c}
\hline
\multirow{2}{*}{System} &  CROHME 14 & CROHME 16 & CROHME 19\\
&$\rm Exp.Rate_{latex}$&$\rm Exp.Rate_{latex}$&$\rm Exp.Rate_{latex}$\\
\hline

WAP~\cite{DBLP:journals/pr/ZhangDZLHHWD17}~ & 40.4 & 37.1 &37.1\\
DenseWAP~\cite{DBLP:conf/icpr/ZhangDD18}~ & 43.0 & 40.1 &41.7\\
PGS~\cite{DBLP:journals/prl/LeAd19}~ &48.78&45.60&-\\
DenseWAP-TD~\cite{DBLP:conf/icml/ZhangDYSW020}~ & 49.1 & 48.5 &51.4\\
PAL~\cite{DBLP:journals/ijcv/WuYZZL20}~ &39.66&49.61&-\\
PAL-v2~\cite{DBLP:conf/cvpr/Le20}~ &48.88&51.53&-\\
WS-WAP~\cite{DBLP:conf/icfhr/TruongNPN20}~ &\textbf{53.65}&\textbf{51.96}&-\\
\textbf{SS-TD} & 52.48 & 51.29 &\textbf{54.32}\\
\hline
\end{tabular}
\end{center}
\end{table}
The comparison among our model and the early algorithms on CROHME 14/16/19 is listed in the Table 4. All the experiment results in Table 4 are selected from published papers. Our model is implemented without using data enhancement and beam search strategies. Obviously, the $\rm Exp.Rate_{latex}$ of SS-TD is 52.48$\rm \%$ on CROHME 2014, 55.29$\rm \%$ on CROHME 2016, and 54.32$\%$ on CROHME 2019. Specifically, compared with the baseline model DenseWAP-TD, our model is 3.33$\%$ better on CROHME 2014 and 6.68$\%$ better on CROHME 2016 than DenseWAP-TD which demonstrates the effectiveness of our improvement.

However, the competitive performance of our method is still lower than some string decoder-based methods, such as~\cite{DBLP:conf/icdar/TruongTN21}~and~\cite{DBLP:conf/icdar/DingHS21}~. The former uses an additional weakly supervised branch to achieve 53.35$\%$ on CROHME 14, the latter uses large scale input images from data augmentation and beam search decoding to achieve 57.72$\%$/61.38$\%$ ExpRate on CROHME 16/19 which take longer to decode iteratively. Indeed, our method pursues the expression spatial structure resolution too much and loses part of the recognition performance of the symbols, which is the main problem that we want to break through in the future. 

\section{Conclusion}
In this paper, we proposed a new tree decoder model integrating spatial attention and syntax rules, which not only improve the structure prediction accuracy, but also decrease the apparent grammar error of mathematical expressions. We illustrate the great performance of SS-TD for offline HMER through the ablation study and comparisons with the advanced methods on CROHME datasets. In the future work, we will further improve the generalization and performance of SS-TD.

\section*{Acknowledgement}This work has been supported by the National Natural Science Foundation of China (No.62176093, 61673182), the Key Realm Research and Development Program of Guangzhou (No.202206030001), and the GuangDong Basic and Applied Basic Research Foundation (No.2021A1515012282).


%
%

\bibliographystyle{splncs04}
\bibliography{ref}





\end{document}